\documentclass[runningheads]{llncs}
\usepackage[T1]{fontenc}
\usepackage{graphicx}

\usepackage{subdef}
\usepackage[ruled,linesnumbered]{algorithm2e}
\usepackage{multirow, multicol}
\usepackage{xcolor}
\definecolor{lightgreen}{rgb}{.80,1,.80}
\definecolor{cobaltblue}{rgb}{.8,.8,1}
\usepackage{soul}
\usepackage[most]{tcolorbox}
\usepackage{subcaption}
\usepackage{array}
\usepackage{makecell}
\usepackage{orcidlink}
\hypersetup{
  colorlinks   = true, 
  urlcolor     = blue, 
  linkcolor    = blue, 
  citecolor   = blue 
}

\begin{document}
\title{EMBLEM: Enhancing Multi-script Table Detection through Masking}

\titlerunning{EMBLEM: Enhancing Multi-script Table Detection through Masking}



\author{
Dhruv Kudale\inst{1}\orcidlink{0000-0002-3862-819X}
\and
Udhay Brahmi\inst{1}\orcidlink{0009-0007-6117-4759}
\and
Ganesh Ramakrishnan\inst{1,2}\orcidlink{0000-0003-4533-2490}
}
\authorrunning{D. Kudale et al.}

\institute{
Indian Institute of Technology Bombay, Mumbai, India 
\email{\{dhruvk,udhaybrahmi,ganesh\}@cse.iitb.ac.in}\\[2pt]
\and
BharatGen, Mumbai, India\\
}

%
            
\maketitle  
\begingroup
\renewcommand{\thefootnote}{}
\footnotetext{D. Kudale and U. Brahmi—These authors contributed equally to this work.}
\footnotetext{D. Kudale—Work done while pursuing MS by Research at IIT Bombay.}
\endgroup

\begin{abstract}
Table detection is a core task in document analysis, supporting downstream applications such as information retrieval, document reconstruction, and visual question answering. Existing deep learning models excel on English and Chinese documents but struggle with multilingual, multi-script documents due to script diversity and the limited availability of labeled data. To address this, we introduce \dataset{} (Multi-script Annotated Documents for Table Detection), a manually curated dataset of $2,323$ table-containing pages spanning 18 languages and 15 scripts, covering diverse domains. We also propose \name, a masking-based paradigm for Multi-script Table Detection (\mtd). \name{} generates masked images that conceal script- and font-specific details, enabling models pre-trained on abundant English documents to focus on script-agnostic page layout. Experiments across three table detection architectures show that \name{} consistently outperforms strong baselines on \dataset{} while remaining competitive on five standard English-dominant benchmarks. Using English-only masked images for fine-tuning and no multi-script training data, \name{} achieves an absolute 20.8\% F1-score gain on \dataset{}. We release \dataset{} and accompanying code and models at \url{https://github.com/IITB-LEAP-OCR/EMBLEM.git}.
\end{abstract}

\section{Introduction}
\label{sec:intro}
\label{sec:intro}
\begin{figure*}[t]
\centering
\frame{\includegraphics[width=0.95\textwidth]{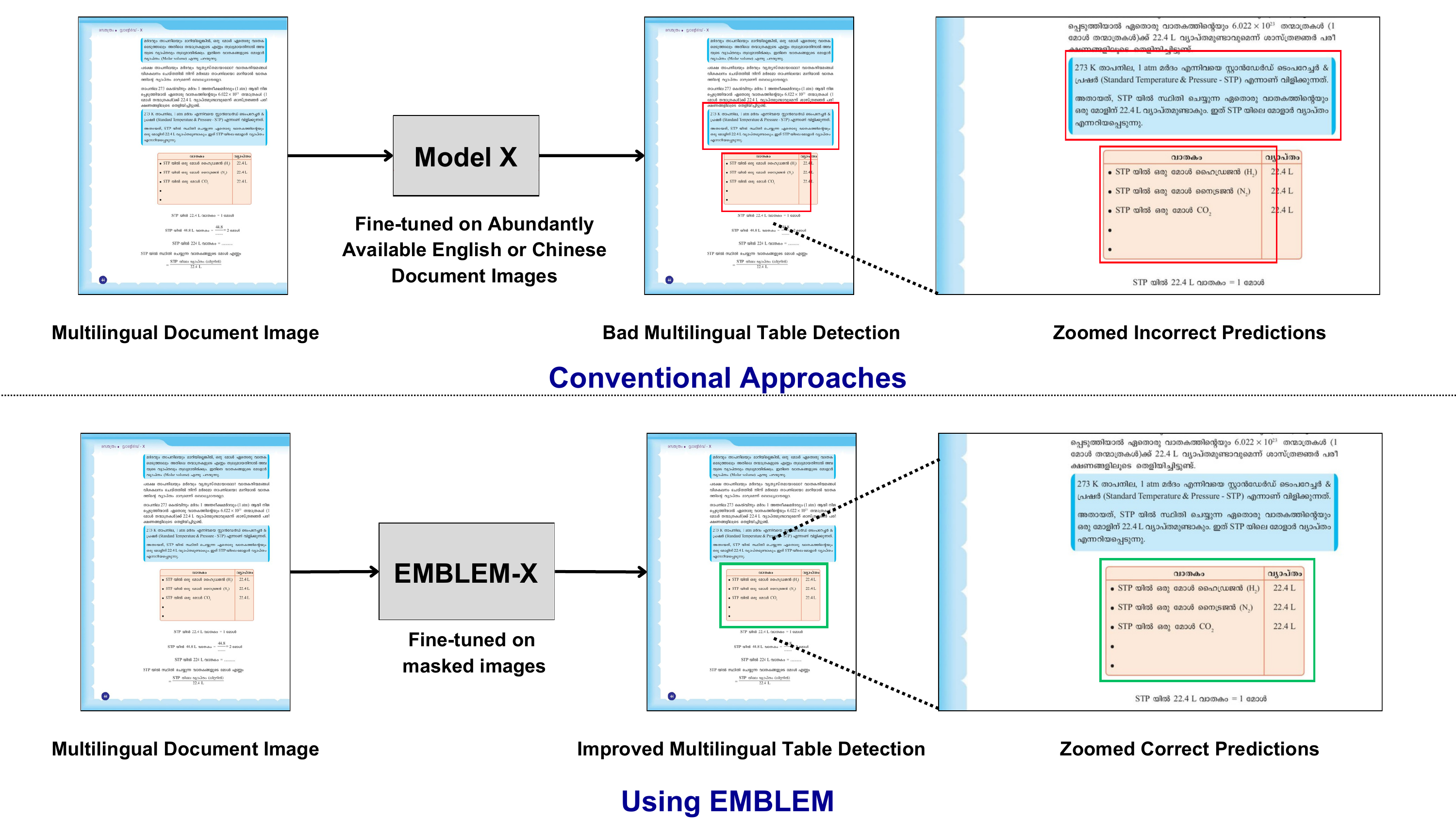}}
\caption{Comparative analysis of conventional \td{} approaches and improvement brought through our method, \name{}, on a Malayalam document page.}
\label{fig:teaser}
\end{figure*}

Table Detection (\td) involves identifying table boundaries in document images, typically using bounding boxes. It is a foundational step in document understanding, supporting downstream tasks such as question answering and structured data extraction. However, \td{} remains challenging due to the wide variety of document formats, writing styles, and, in particular, the presence of multiple scripts. While an ideal \td{} model should work across scripts and languages, most existing research has focused on English or Chinese, where datasets and models are readily available. As illustrated in Fig.~\ref{fig:teaser}, models trained on such monolingual data often fail to generalize to documents in other scripts. Even a change in script alone, without altering layout or document structure, can cause a substantial drop in detection accuracy, often leading to missed or fragmented table predictions. This highlights the largely unexplored area of \mtd{}, where detecting tables in documents with diverse scripts and languages remains a significant challenge. The scarcity of annotated multi-script datasets further exacerbates the problem, as most existing resources are monolingual or bilingual and limited to specific domains, making the training of robust \mtd{} models both difficult and resource-intensive. To address these challenges, we make three key contributions, summarized as follows:

\begin{enumerate}
\item We release \textbf{\dataset}, a manually annotated multi-script table detection benchmark with 2,323 pages covering 15 scripts, 18 languages, and diverse document domains.
\item We propose \textbf{\name}, an image masking-based finetuning paradigm that learns script-agnostic structural cues, improving \mtd{} performance.
\item We demonstrate a suite of models leveraging \name, achieving strong cross-script performance on \dataset{} even when trained only on masked, abundantly labeled English or Chinese documents, while maintaining competitive performance on standard English-dominated benchmarks.
\end{enumerate}

\section{Related Work}

\textbf{\td{} Approaches:} Earlier computer vision (CV) based methods \cite{tintin,pattern-recog,white-space} and machine learning (ML) based methods \cite{ml-classifiers,ml-random-forest,ml-svm} relied on heuristics or hand-crafted features such as alignment, spacing, and ruled lines, but were brittle on complex layouts. Deep learning (DL) approaches now dominate this field, treating \td{} as an object detection task with architectures including Faster R-CNN \cite{faster-rcnn,frcnn-deepdesrt}, Mask R-CNN \cite{mask-rcnn,layoutparser}, Cascade Mask R-CNN \cite{cascadTabNet}, YOLO \cite{yolo,yolo-ptn,yolo-td}, SSD \cite{ssd-td}, and transformer-based approaches such as DETR \cite{detr,sam-detr} and TATR \cite{tatr-pub-1m}. Some methods jointly address \td{} and structure recognition \cite{fintabnet,multi-type-td-tsr}, while vision-language models show promise \cite{doclayout,ld-doc,tabpedia,udop,layoutlmv3,dit} but face challenges in cross-script scenarios due to language-specific pretraining.
\\
\textbf{\td{} Datasets:} Existing benchmarks are largely English-centric and limited to Latin-script documents. Popular datasets include ICDAR 2013 \cite{icdar13}, ICDAR 2017 \cite{icdar17}, ICDAR 2019 \ctdar{} \cite{icdar19}, UNLV \cite{unlv}, UW3 \cite{uw3}, Marmot \cite{marmot}, TableBank \cite{tablebank}, \pubm{} \cite{tatr-pub-1m}, IIIT-AR-13K \cite{iiit-ar}, PubLayNet \cite{publaynet}, and DocLayNet \cite{doclaynet} (Table~\ref{tab:td-datasets}). While they have enabled significant progress in \td{}, their coverage of scripts, non-Latin writing systems, and domains is limited. Consequently, models trained on these datasets tend to perform poorly when applied to multi-script documents or cross-domain scenarios, highlighting the need for benchmarks and evaluation protocols that test multi-script and domain-robust \td{} approaches.
\begin{table*}[t]
\centering
\small
\setlength{\tabcolsep}{4pt}
\renewcommand{\arraystretch}{1}
\caption{Comparison of existing benchmarks by test-set size, script coverage, and document domains. Table-specific benchmarks have only tables annotated whereas other benchmarks focus on different document layout classes as well.}
\begin{tabular}{|
>{\centering\arraybackslash}p{2.5cm}|
>{\centering\arraybackslash}p{1cm}|
>{\centering\arraybackslash}p{3.4cm}|
>{\centering\arraybackslash}p{2.7cm}|
>{\centering\arraybackslash}p{1.2cm}|}
\hline
\textbf{Dataset} & \textbf{Size (Test)} & \textbf{Scripts Covered} & \textbf{Domains} & \textbf{Table-Specific} \\
\hline
ICDAR 2013 \cite{icdar13} & 233 & Latin & \multirow{8}{*}{\centering Scientific} & Yes \\
\cline{1-3}\cline{5-5}
ICDAR 2017 \cite{icdar17} & 817 & Latin &  & Yes \\
\cline{1-3}\cline{5-5}
\ctdar{} Track A Modern \cite{icdar19} & 240 & Latin, Chinese &  & Yes \\
\cline{1-3}\cline{5-5}
UNLV \cite{unlv} & 2.9K & Latin &  & No \\
\cline{1-3}\cline{5-5}
UW3 \cite{uw3} & 1.6K & Latin &  & Yes \\
\cline{1-3}\cline{5-5}
Marmot \cite{marmot} & 2K & Latin, Chinese &  & Yes \\
\cline{1-3}\cline{5-5}
TableBank \cite{tablebank} & 8K & Latin &  & Yes \\
\cline{1-3}\cline{5-5}
\pubm{} \cite{tatr-pub-1m} & 92K & Latin &  & Yes \\
\cline{1-3}\cline{5-5}
PubLayNet \cite{publaynet} & 3.9K & Latin &  & No \\
\hline
STDW \cite{stdw} & 7K & Latin, Japanese, Devanagari &
\multirow{4}{*}{\makecell[c]{Commercial, \\ Administrative}} & Yes \\
\cline{1-3}\cline{5-5}
DocLayNet \cite{doclaynet} & 2.2K & Latin &  & No \\
\cline{1-3}\cline{5-5}
IIIT-AR-13K \cite{iiit-ar} & 2.1K & Latin &  & No \\
\cline{1-3}\cline{5-5}
TNCR \cite{tncr} & 1K & Predominantly Latin &  & Yes \\
\hline
\textbf{\center MANDALA (Ours)} &
\textbf{\center 2.3K} &
\textbf{Bengali, Cyrillic, Devanagari, Greek, Gujarati, Gurmukhi, Japanese, Kannada, Khmer, Malayalam, Odia, Perso-Arabic, Tamil, Telugu, Thai} &
\textbf{\small Scientific, Linguistic, Social, Commercial, Administrative, Literature, Philosophy} &
\textbf{\center Yes} \\
\hline
\end{tabular}
\label{tab:td-datasets}
\end{table*}

\noindent
\textbf{Towards Script-Agnostic Generalization:} Annotating tables across scripts is costly, motivating data-efficient approaches such as transfer learning, domain adaptation \cite{close-domain-fine-tuning,crossdod}, hybrid models \cite{hybridtabnet}, and semi-supervised learning \cite{semi-sup-td}, which remain English-centric and largely ignore script variation. Table structure is largely script-independent, defined by ruled lines, grids, and whitespace, enabling the possibility of cross-script generalization via existing English-labeled data. Masking-based augmentations have been shown to improve generalization in object detection \cite{mask-aug-cnn,mask-aug-hidenseek}, but their effectiveness for cross-script \td{} is underexplored. Motivated by this, we propose \name{}, a masking-based fine-tuning strategy leveraging script-agnostic cues, improving \mtd{} performance while remaining competitive on English benchmarks. To support systematic \mtd{} evaluation, we introduce \dataset{}, a benchmark described in the upcoming Section~\ref{sec:mandala}.


\section{\textbf{\dataset{}} Dataset}
\label{sec:mandala}

We introduce \dataset, a multilingual, multi-script, and multi-domain document dataset designed to evaluate \mtd{} under realistic and diverse conditions. \dataset{} is designed primarily as a test and analysis benchmark, rather than a large-scale training corpus. In contrast to existing benchmarks that are predominantly English-centric or restricted to a limited set of scripts or domains, \dataset{} explicitly varies along two axes: script and domain.

\textbf{Script and Language Diversity:} \dataset{} contains 2,323 document images spanning 18 languages written in 15 distinct scripts, covering Indic, European, Middle Eastern, and Southeast Asian languages. The dataset explicitly disentangles language and script, with multiple languages sharing the same script (e.g., Hindi and Marathi in Devanagari; Persian and Urdu in Perso-Arabic), enabling targeted evaluation of script-level generalization. A balanced yet realistic distribution is maintained, with per-language sample counts ranging from fewer than 100 to over 200 images (Table~\ref{tab:mandala-stats}), ensuring meaningful representation of low-resource languages and scripts.

\begin{table}[]
\caption{Distribution of different scripts (and languages) in \dataset{}.}
\setlength{\tabcolsep}{6pt}
\renewcommand{\arraystretch}{1}
\centering
\begin{tabular}{|c|c|c|c|}
\hline
\textbf{Language} & \textbf{Pages} & \textbf{Script} & \textbf{Domain} \\ \hline
Bengali & 164 & Bengali & Scientific \\ \hline
Greek & 104 & Greek & Linguistic \\ \hline
Gujarati & 118 & Gujarati & Scientific \\ \hline
Hindi & 186 & Devanagari & Administrative, Commercial \\ \hline
Japanese & 124 & Japanese & Philosophy \\ \hline
Kannada & 99 & Kannada & Social, Philosophy \\ \hline
Khmer & 206 & Khmer & Linguistic \\ \hline
Malayalam & 105 & Malayalam & Philosophy, Commercial \\ \hline
Marathi & 134 & Devanagari & Commercial \\ \hline
Odia & 125 & Odia & Scientific, Social \\ \hline
Persian & 133 & Perso-Arabic & Linguistic \\ \hline
Punjabi & 125 & Gurmukhi & Scientific, Philosophy, Social \\ \hline
Russian & 172 & Cyrillic & Linguistic \\ \hline
Tamil & 93 & Tamil & Literature, Philosophy \\ \hline
Telugu & 110 & Telugu & Literature, Scientific \\ \hline
Thai & 129 & Thai & Social \\ \hline
Ukrainian & 94 & Cyrillic & Linguistic \\ \hline
Urdu & 102 & Perso-Arabic & Scientific \\ \hline
\textbf{Overall} & \textbf{2323} &  &  \\ \hline
\end{tabular}
\label{tab:mandala-stats}
\end{table}

\textbf{Domain Diversity:} \dataset{} spans seven coarse-grained domains, including \textbf{Scientific} textbooks (mathematics, physics, chemistry, biology with tables, equations and figures), \textbf{Linguistic} language learning materials (grammar workbooks with conjugation tables), \textbf{Social} studies content (history, geography, civics with timelines and illustrative tables), \textbf{Commercial} documents (business, economics, finance with numerical tables), \textbf{Administrative} materials (circulars, schedules, notices from regulatory institutions), \textbf{Literature} ( narrative text with occasional structured tabular elements), and \textbf{Philosophy} (value education, general knowledge and pictorial representation). Each language subset spans one or more domains, with overall distribution to have approximately 80\% single-column and 20\% two-column layouts. Each document is assigned a single dominant domain.

\textbf{\dataset{} Annotation}
All document images in \dataset{} were manually annotated following strict guidelines to ensure high-quality ground truth, with a focus on table regions and structural variations. Tables are treated as a single class, and annotation follows the case-wise rules below:

\begin{itemize}
\item \textbf{Table Types:} All \textit{bordered}, \textit{partially bordered}, and \textit{unbordered} tables are annotated as tables, as a single class.
\item \textbf{Unbordered Tables:} Identified based on regular grid alignment, consistent inter-cell spacing, or equidistant multi-cell layouts.
\item \textbf{Rotated and Skewed Tables:} Tables rotated by $90^\circ$ (left or right), as well as tables appearing in rotated or skewed pages, are annotated.
\item \textbf{Tables of Contents:} are annotated as a single contiguous table on a page.
\item \textbf{Large Tables:} A large table with uniform and continuous boundaries is annotated as a single table. If a large table contains multiple sub-tables separated by subtitles or headers lying outside the table boundary, each sub-table is annotated independently.
\item \textbf{Table Captions:} are included only when they lie entirely within the table boundary, else not engulfed in the bounding box.
\item \textbf{Legends and Indices:} Legends in maps or figures are annotated.
\item \textbf{Lists and Pseudo-Tables:} Only completely or partially bordered numbered lists with explicit column demarcation are annotated as tables; plain text lists are excluded.
\item \textbf{Nested Tables:} are extremely rare and are not annotated separately.
\end{itemize}

\begin{figure}[h]
     \centering
    \frame{\includegraphics[width=0.8\textwidth]{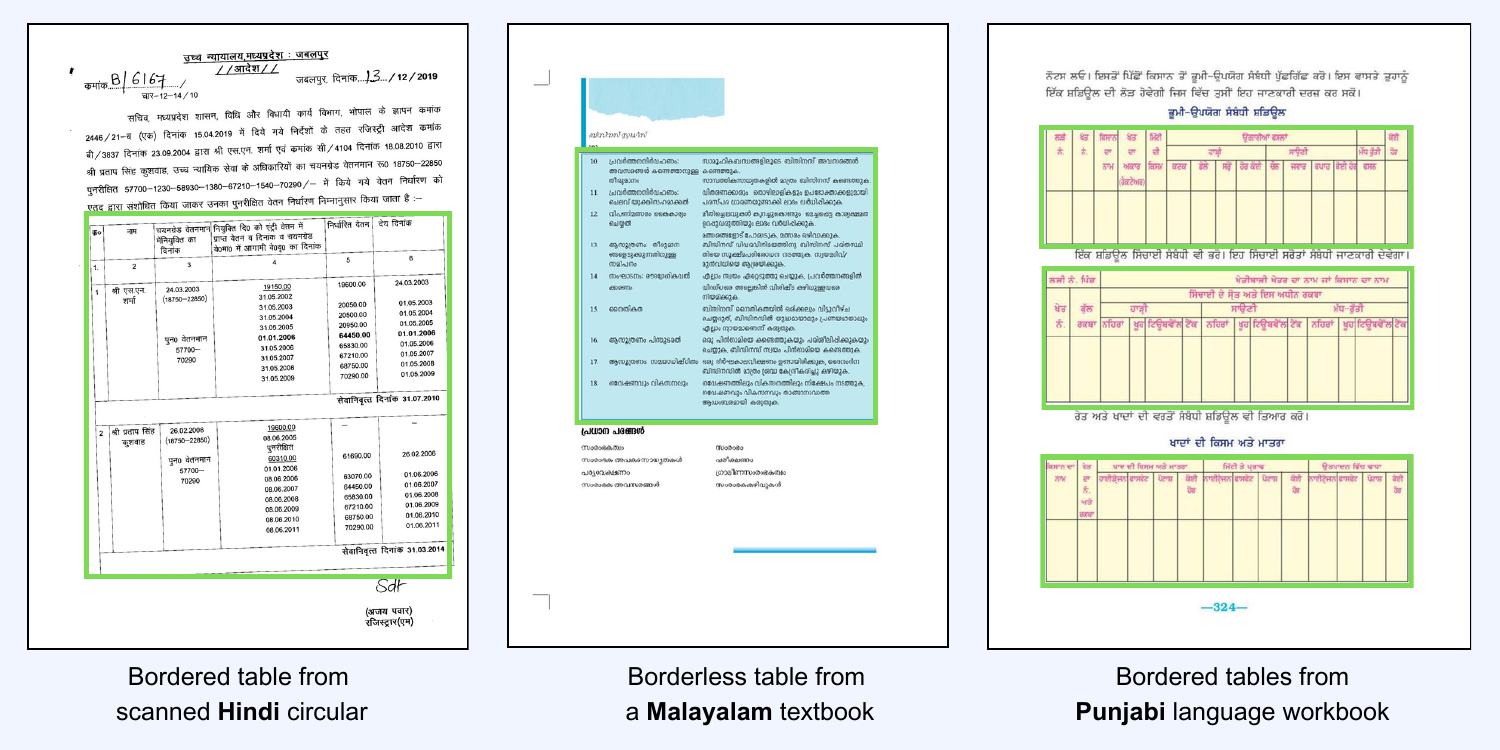}}
    \caption{Sample images from \dataset{} with tables in bounding boxes.}
    \label{fig:mtd}
\end{figure}

All annotations were manually curated and independently reviewed before final inclusion. \dataset{} spans 18 languages, 15 scripts, and 7 document domains. The source documents used in \dataset{} were obtained from publicly accessible resources, and source metadata is provided to facilitate attribution. A few examples from \dataset{} are shown in Fig.~\ref{fig:mtd}.

\section{Experiments}
\label{sec:experiments}

This section presents the experimental framework used to evaluate \name. \name{} is implemented as a masking-based fine-tuning paradigm that suppresses script-dependent visual cues. We describe the \name{} masking process, dataset construction, and experimental protocol used across all evaluations.

\subsection{Masking using \name}
\label{sec:paradigm-masking}
\begin{figure}[t]
    \centering
    \frame{\includegraphics[width=0.8\textwidth]{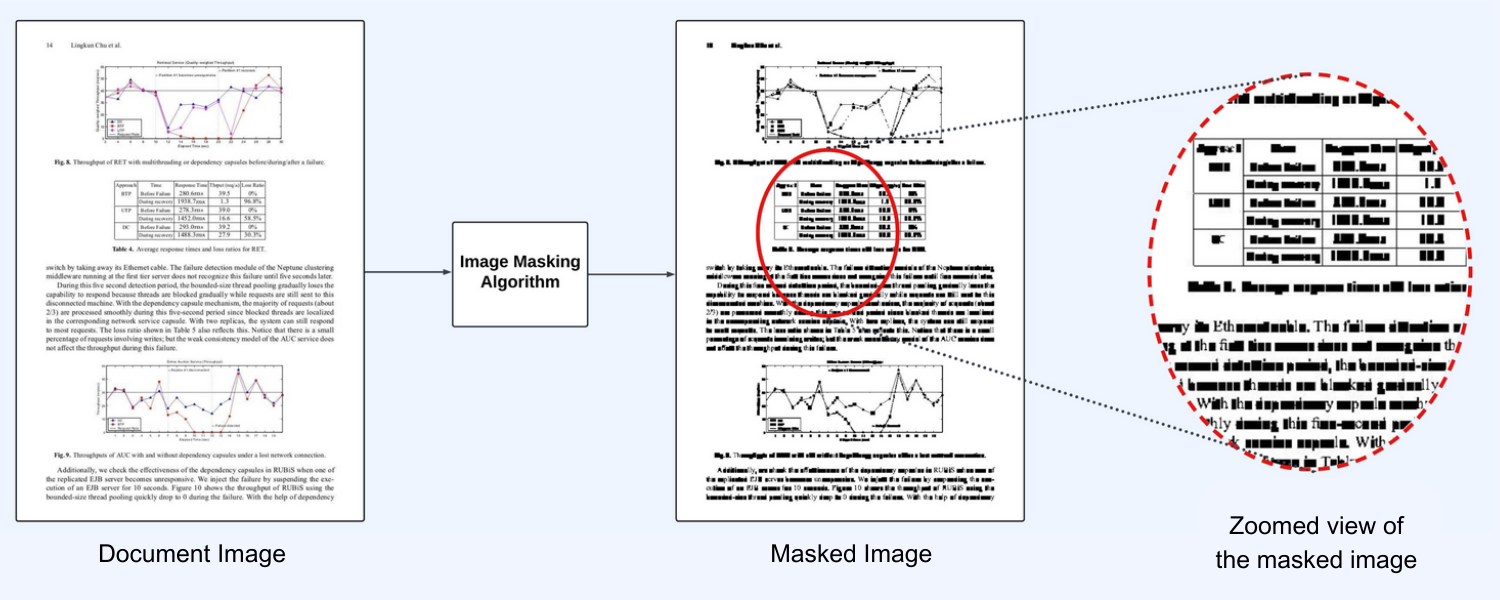}}
    \caption{Masking of a \ctdar{} image illustrating the strategy used in \name{}.}
    \label{fig:mask}
\end{figure}

The objective of \name{} is to enable \td{} models to generalize across diverse document scripts and domains without requiring multi-script annotations. To this end, \name{} suppresses script-specific visual cues while preserving script-agnostic structural information such as whitespace, ruled lines, alignment, and table boundaries thereby encouraging models to rely on layout rather than script. By masking fine-grained text strokes while retaining layout primitives, \name{} shifts learning from appearance-driven cues to structural regularities that are invariant across scripts. Building on the strong layout priors encoded in \td{} models pretrained on English documents, \name{} fine-tunes these models on masked document images (Fig.~\ref{fig:mask}) to mitigate script bias while maintaining sensitivity to document structure.

Unlike generic masking-based augmentations, \name{} explicitly suppresses fine-grained, script-dependent stroke patterns while preserving layout primitives essential for table localization, including lines, whitespace, and alignment. This design is critical to remove linguistic appearance cues without distorting document structure. Unlike OCR-based masking, which operates at the character, word, or line level and often over-masks large rectangular regions, distorting layout cues and introducing visual artifacts, \name{} employs a pixel-level, CV-based masking strategy. Specifically, the input image is converted to grayscale and binarized using Otsu’s thresholding~\cite{otsu}. Contours are extracted from the inverted binary image and enclosed within bounding boxes. Bounding boxes smaller than a configurable threshold defined as a fraction of the image dimensions are masked using small filled rectangles. This selectively suppresses script-specific strokes while minimizing interference with layout elements such as table borders and cell lines. The complete procedure is summarized in Algorithm \ref{alg:masking}.

\begin{algorithm}[h]
\caption{Image Masking Algorithm used in \name{}}
\label{alg:masking}
\small
\SetKwInOut{Input}{Input}
\SetKwInOut{Output}{Output}
\Input{\textit{image}, \textit{threshold}}
\Output{\textit{masked\_image}}

\textit{gray\_image} $\gets$ \textsc{ConvertToGrayscale}(\textit{image})\;
\textit{binary\_image} $\gets$ \textsc{OtsuThresholding}(\textit{gray\_image})\;
\textit{inverted\_binary} $\gets$ \textsc{InvertImage}(\textit{binary\_image})\;
\textit{width}, \textit{height} $\gets$ \textsc{ImageDimensions}(\textit{image})\;
\textit{contours} $\gets$ \textsc{FindContours}(\textit{inverted\_binary})\;
\textit{masked\_image} $\gets$ \textit{binary\_image}\;
\ForEach{\textit{contour} \textbf{in} \textit{contours}}{
    \textit{x, y, w, h} $\gets$ \textsc{GetBoundingRect}(\textit{contour})\;
    \If{\textit{h} $<$ \textit{threshold * height} \textbf{and} \textit{w} $<$ \textit{threshold * width}}{
        \textsc{DrawRectangle}(\textit{masked\_image}, \textit{x, y, w, h})\;
    }
}
\Return \textit{masked\_image}
\end{algorithm}

\subsection{Datasets}
\label{sec:datasets}

\textbf{Masked Dataset for Fine-tuning:}
The masked dataset used for fine-tuning \name{} models is generated using Algorithm~\ref{alg:masking}. Source images are obtained from the Kaggle Synthetic Table Detection~\footnote{Synthetic Table Detection \url{https://www.kaggle.com/datasets/champ333999/synthetic-image-data-for-table-detection}} and General Table Detection~\footnote{\url{https://www.kaggle.com/datasets/rhtsingh/general-table-recognition-dataset}} datasets, the latter being a superset of Marmot~\cite{marmot}, the GitHub \td{} dataset~\cite{table-github}, and the ICDAR 2019 \ctdar{} training set~\cite{icdar19}. The dataset consists predominantly of labeled English document pages, with a small fraction of Chinese pages (around $10\%$); no other languages are included. As a result, \td{} models are not exposed to multi-script document images at any stage of fine-tuning. After masking, images are randomly split into $2{,}000$ training and $183$ validation samples, with original image resolutions and table bounding box annotations preserved. This strategy allows the reuse of existing labeled document pages to obtain a language-agnostic representation without requiring additional multi-script annotations. To determine the masking granularity, we experiment with multiple height and width thresholds defined relative to the image dimensions. Fine-tuning is performed using a \yolov{} model pretrained on the \pubm{} dataset. Threshold selection is guided by performance on the bilingual \ctdar{} test set. As determined by our empirical analysis, a threshold of $1\%$ achieves the best trade-off between suppressing script-specific cues and preserving structural alignment. Lower thresholds fail to sufficiently mask textual content, while higher thresholds introduce visual distortions. Consequently, a $1\%$ threshold is used to generate the final masked dataset and \name{} finetuning.
\\
\textbf{Evaluation Datasets:} We evaluate models fine-tuned using \name{} on six \td{} benchmarks: \ctdar{} (bilingual English–Chinese, for cross-script robustness), ICDAR 2013 (standard English \td{}), \pubm{} (large-scale scientific documents), IIIT-AR-13K (Latin script administrative documents in 4 languages), TNCR (complex real-world table layouts), and eventually on \dataset{}.

\subsection{Fine-tuning \textbf{\name{}} Models}
\label{sec:ftn}
We fine-tune three widely used object detection–based \td{} architectures: \yolov{}, \tatr{}, and DocLayout (\docl{}) using the masked images described in Section \ref{sec:datasets}, to highlight the model-agnostic applicability of \name{}. These architectures were selected for their strong \td{} performance on English documents. All experiments were conducted on an NVIDIA A6000 GPU.
The \textbf{\name-\yolov{}} model, pretrained on the \pubm{} dataset \cite{yolo-pub}, was fine-tuned for 15 epochs with a batch size of 8 and an input resolution of $640 \times 640$. For \textbf{\name-\tatr{}}, we fine-tuned the DETR-based model with a ResNet-18 backbone \cite{detr,resnet}, also pretrained on \pubm{} \cite{tatr-pub-1m}, for 30 epochs with a batch size of 4, retaining the default table detection configuration \cite{tatr-pub-1m}. Finally, we fine-tuned the YOLOv10-based \textbf{\name-\docl{}} model \cite{doclayout}, pretrained on DocSynth-300K and DocStructBench \cite{doclayout}, for 200 epochs with a batch size of 32 on masked images resized to 640 pixels on the longest side. Default Ultralytics training configurations\footnote{Ultralytics \url{https://www.ultralytics.com}}
 were used for fine-tuning \yolov{} and \docl{} models.

\section{Results}
We provide a comparative analysis of our \name{} models, presenting their detection scores across various test sets against different baselines.
\label{sec:results}

\subsection{Performance Gain for \textbf{\mtd{}}}
\begin{figure}[h]
     \centering
    \frame{\includegraphics[width =0.9\textwidth]{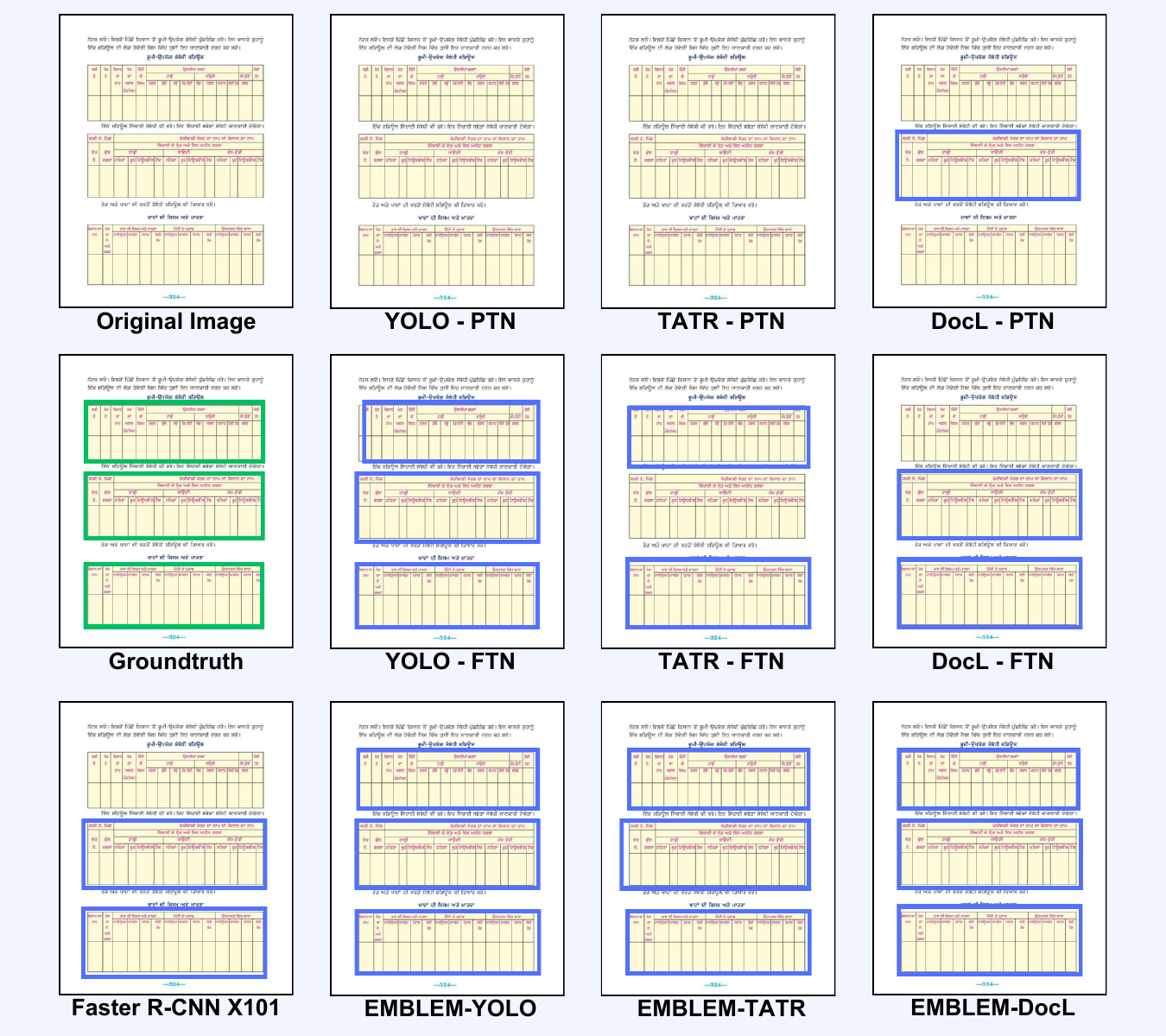}}
    \caption{Results on a sample Punjabi page from \dataset.}
    \label{fig:qualitative}
\end{figure}

This section highlights the performance of \name{} relative to its pretrained counterparts (\yolo{}-PTN, \tatr{}-PTN, and \docl{}-PTN). To isolate the impact of masking, we also report baselines where these models are fine-tuned on the original, unmasked images using the same training settings described in Section~\ref{sec:ftn} (\yolo{}-FTN, \tatr{}-FTN, and \docl{}-FTN). All models are evaluated using a confidence threshold of 0.75 and a Non-Maximum Suppression IoU threshold of 0.1. As shown in Table~\ref{tab:perf-impr}, \name{} consistently outperforms both PTN and FTN baselines on all benchmarks. In particular, \name-\yolov{} achieves the best performance on \ctdar{} \cite{icdar19}, while other \name{} variants lead on the remaining datasets. These improvements demonstrate the effectiveness of \name{} in handling multi-script content and complex document layouts. Notably, \name-\docl{} also exhibits consistent gains, indicating that the benefits of \name{} are independent of the underlying pretraining corpus: \pubm{} for the \yolov{} and \tatr{} variants, and DocSynth-300K and DocStructBench for the \docl{} variants. Fig.~\ref{fig:qualitative} provides qualitative comparisons on a representative page from \dataset{}, further illustrating the robustness of \name{} across scripts and layouts.

\begin{table}[]
\setlength{\tabcolsep}{6pt}
\renewcommand{\arraystretch}{1.1}
\centering
\caption{Performance comparison at IoU 0.5 and 0.7. Best scores indicate that \name{} consistently outperforms all baselines across all four datasets.}
\begin{tabular}{|cc|ccc|ccc|}
\hline
\multicolumn{2}{|c|}{\textbf{IOU Threshold}} & \multicolumn{3}{c|}{\textbf{0.5}} & \multicolumn{3}{c|}{\textbf{0.7}} \\ \hline
\multicolumn{1}{|c|}{\textbf{Dataset}} & \textbf{Model} & \multicolumn{1}{c|}{\textbf{P}} & \multicolumn{1}{c|}{\textbf{R}} & \textbf{F} & \multicolumn{1}{c|}{\textbf{P}} & \multicolumn{1}{c|}{\textbf{R}} & \textbf{F} \\ \hline
\multicolumn{1}{|c|}{\multirow{9}{*}{\begin{tabular}[c]{@{}c@{}}\textbf{ICDAR 13}            \\ \textbf{(1 language,}\\ \textbf{1 script)}\end{tabular}}} & \yolov-PTN & \multicolumn{1}{c|}{96.9} & \multicolumn{1}{c|}{95.4} & 96.2 & \multicolumn{1}{c|}{96.9} & \multicolumn{1}{c|}{95.4} & 96.2 \\  
\multicolumn{1}{|c|}{} & \tatr-PTN & \multicolumn{1}{c|}{97.7} & \multicolumn{1}{c|}{96.2} & 96.9 & \multicolumn{1}{c|}{97.7} & \multicolumn{1}{c|}{96.2} & 96.9 \\  
\multicolumn{1}{|c|}{} & \docl-PTN & \multicolumn{1}{c|}{95.3} & \multicolumn{1}{c|}{94.8} & 95.1 & \multicolumn{1}{c|}{93.8} & \multicolumn{1}{c|}{93.2} & 93.5 \\  
\multicolumn{1}{|c|}{} & \yolov-FTN & \multicolumn{1}{c|}{98.1} & \multicolumn{1}{c|}{\hlg{99.2}} & 98.6 & \multicolumn{1}{c|}{94.9} & \multicolumn{1}{c|}{96.1} & 95.5 \\  
\multicolumn{1}{|c|}{} & \tatr-FTN & \multicolumn{1}{c|}{\hlg{99.2}} & \multicolumn{1}{c|}{\hlg{99.2}} & \hlg{99.2} & \multicolumn{1}{c|}{\hlg{98.4}} & \multicolumn{1}{c|}{\hlg{98.4}} & \hlg{98.4} \\  
\multicolumn{1}{|c|}{} & \docl-FTN & \multicolumn{1}{c|}{98.4} & \multicolumn{1}{c|}{98.2} & 98.3 & \multicolumn{1}{c|}{96.1} & \multicolumn{1}{c|}{95.8} & 96.0 \\  
\multicolumn{1}{|c|}{} & \textbf{\name-\yolo} & \multicolumn{1}{c|}{97.0} & \multicolumn{1}{c|}{98.8} & 97.9 & \multicolumn{1}{c|}{95.8} & \multicolumn{1}{c|}{97.3} & 96.5 \\  
\multicolumn{1}{|c|}{} & \textbf{\name-\tatr} & \multicolumn{1}{c|}{98.8} & \multicolumn{1}{c|}{\hlg{99.2}} & 99.0 & \multicolumn{1}{c|}{96.1} & \multicolumn{1}{c|}{96.5} & 96.3 \\  
\multicolumn{1}{|c|}{} & \textbf{\name-\docl} & \multicolumn{1}{c|}{96.2} & \multicolumn{1}{c|}{96.9} & 96.6 & \multicolumn{1}{c|}{93.9} & \multicolumn{1}{c|}{94.5} & 94.2 \\ \hline
\multicolumn{1}{|c|}{\multirow{9}{*}{\begin{tabular}[c]{@{}c@{}}\textbf{\ctdar}\\ \textbf{(2 languages,} \\ \textbf{2 scripts)}\end{tabular}}} & \yolov-PTN & \multicolumn{1}{c|}{87.4} & \multicolumn{1}{c|}{82.9} & 85.1 & \multicolumn{1}{c|}{86.0} & \multicolumn{1}{c|}{81.6} & 83.8 \\  
\multicolumn{1}{|c|}{} & \tatr-PTN & \multicolumn{1}{c|}{80.1} & \multicolumn{1}{c|}{73.1} & 76.4 & \multicolumn{1}{c|}{76.2} & \multicolumn{1}{c|}{69.9} & 72.9 \\  
\multicolumn{1}{|c|}{} & \docl-PTN & \multicolumn{1}{c|}{86.0} & \multicolumn{1}{c|}{83.7} & 84.8 & \multicolumn{1}{c|}{82.9} & \multicolumn{1}{c|}{80.8} & 81.8 \\  
\multicolumn{1}{|c|}{} & \yolov-FTN & \multicolumn{1}{c|}{97.8} & \multicolumn{1}{c|}{98.2} & 98.0 & \multicolumn{1}{c|}{97.3} & \multicolumn{1}{c|}{97.7} & 97.5 \\  
\multicolumn{1}{|c|}{} & \tatr-FTN & \multicolumn{1}{c|}{97.9} & \multicolumn{1}{c|}{96.8} & 97.3 & \multicolumn{1}{c|}{97.4} & \multicolumn{1}{c|}{95.1} & 96.3 \\  
\multicolumn{1}{|c|}{} & \docl-FTN & \multicolumn{1}{c|}{\hlg{98.6}} & \multicolumn{1}{c|}{95.3} & 96.9 & \multicolumn{1}{c|}{\hlg{98.2}} & \multicolumn{1}{c|}{94.9} & 96.5 \\  
\multicolumn{1}{|c|}{} & \textbf{\name-\yolo} & \multicolumn{1}{c|}{98.0} & \multicolumn{1}{c|}{\hlg{98.3}} & \hlg{98.1} & \multicolumn{1}{c|}{97.4} & \multicolumn{1}{c|}{\hlg{97.8}} & \hlg{97.6} \\  
\multicolumn{1}{|c|}{} & \textbf{\name-\tatr} & \multicolumn{1}{c|}{95.9} & \multicolumn{1}{c|}{94.1} & 95.0 & \multicolumn{1}{c|}{95.5} & \multicolumn{1}{c|}{93.7} & 94.6 \\  
\multicolumn{1}{|c|}{} & \textbf{\name-\docl} & \multicolumn{1}{c|}{97.7} & \multicolumn{1}{c|}{97.5} & 97.6 & \multicolumn{1}{c|}{97.5} & \multicolumn{1}{c|}{97.1} & 97.3 \\ \hline
\multicolumn{1}{|c|}{\multirow{9}{*}{\textbf{\begin{tabular}[c]{@{}c@{}}IIIT-AR-13K\\ (4 languages,\\ 1 script)\end{tabular}}}} & \yolov-PTN & \multicolumn{1}{c|}{74.1} & \multicolumn{1}{c|}{69.5} & 71.7 & \multicolumn{1}{c|}{68.5} & \multicolumn{1}{c|}{64.8} & 66.6 \\  
\multicolumn{1}{|c|}{} & \tatr-PTN & \multicolumn{1}{c|}{74.8} & \multicolumn{1}{c|}{67.6} & 71.0 & \multicolumn{1}{c|}{65.9} & \multicolumn{1}{c|}{59.9} & 62.8 \\  
\multicolumn{1}{|c|}{} & \docl-PTN & \multicolumn{1}{c|}{83.2} & \multicolumn{1}{c|}{80.6} & 81.9 & \multicolumn{1}{c|}{81.2} & \multicolumn{1}{c|}{79.3} & 80.3 \\  
\multicolumn{1}{|c|}{} & \yolov-FTN & \multicolumn{1}{c|}{91.8} & \multicolumn{1}{c|}{89.8} & 90.8 & \multicolumn{1}{c|}{89.3} & \multicolumn{1}{c|}{87.7} & 88.5 \\  
\multicolumn{1}{|c|}{} & \tatr-FTN & \multicolumn{1}{c|}{94.2} & \multicolumn{1}{c|}{92.9} & 93.6 & \multicolumn{1}{c|}{91.1} & \multicolumn{1}{c|}{90.0} & 90.5 \\  
\multicolumn{1}{|c|}{} & \docl-FTN & \multicolumn{1}{c|}{92.1} & \multicolumn{1}{c|}{89.0} & 90.5 & \multicolumn{1}{c|}{89.6} & \multicolumn{1}{c|}{87.0} & 88.3 \\  
\multicolumn{1}{|c|}{} & \textbf{\name-\yolo} & \multicolumn{1}{c|}{92.8} & \multicolumn{1}{c|}{\hlg{93.3}} & 93.1 & \multicolumn{1}{c|}{90.4} & \multicolumn{1}{c|}{\hlg{91.1}} & 90.7 \\  
\multicolumn{1}{|c|}{} & \textbf{\name-\tatr} & \multicolumn{1}{c|}{\hlg{94.5}} & \multicolumn{1}{c|}{93.0} & \hlg{93.7} & \multicolumn{1}{c|}{90.7} & \multicolumn{1}{c|}{89.6} & 90.1 \\  
\multicolumn{1}{|c|}{} & \textbf{\name-\docl} & \multicolumn{1}{c|}{93.6} & \multicolumn{1}{c|}{93.0} & 93.3 & \multicolumn{1}{c|}{\hlg{91.1}} & \multicolumn{1}{c|}{90.8} & \hlg{91.0} \\ \hline
\multicolumn{1}{|c|}{\multirow{9}{*}{\textbf{\begin{tabular}[c]{@{}c@{}}MANDALA\\ (18 languages,\\ 15 scripts)\end{tabular}}}} & \yolov-PTN & \multicolumn{1}{c|}{55.0} & \multicolumn{1}{c|}{49.8} & 52.5 & \multicolumn{1}{c|}{51.6} & \multicolumn{1}{c|}{46.9} & 49.2 \\  
\multicolumn{1}{|c|}{} & \tatr-PTN & \multicolumn{1}{c|}{55.6} & \multicolumn{1}{c|}{49.2} & 52.2 & \multicolumn{1}{c|}{50.6} & \multicolumn{1}{c|}{45.0} & 47.6 \\  
\multicolumn{1}{|c|}{} & \docl-PTN & \multicolumn{1}{c|}{71.5} & \multicolumn{1}{c|}{67.9} & 69.6 & \multicolumn{1}{c|}{71.0} & \multicolumn{1}{c|}{67.5} & 69.2 \\  
\multicolumn{1}{|c|}{} & \yolov-FTN & \multicolumn{1}{c|}{86.4} & \multicolumn{1}{c|}{83.5} & 84.9 & \multicolumn{1}{c|}{85.5} & \multicolumn{1}{c|}{82.7} & 84.1 \\  
\multicolumn{1}{|c|}{} & \tatr-FTN & \multicolumn{1}{c|}{88.4} & \multicolumn{1}{c|}{83.2} & 85.7 & \multicolumn{1}{c|}{85.9} & \multicolumn{1}{c|}{81.1} & 83.4 \\  
\multicolumn{1}{|c|}{} & \docl-FTN & \multicolumn{1}{c|}{81.7} & \multicolumn{1}{c|}{76.2} & 78.8 & \multicolumn{1}{c|}{80.5} & \multicolumn{1}{c|}{75.3} & 77.8 \\  
\multicolumn{1}{|c|}{} & \textbf{\name-\yolo} & \multicolumn{1}{c|}{87.6} & \multicolumn{1}{c|}{86.9} & 87.3 & \multicolumn{1}{c|}{86.4} & \multicolumn{1}{c|}{85.8} & 86.1 \\  
\multicolumn{1}{|c|}{} & \textbf{\name-\tatr} & \multicolumn{1}{c|}{88.9} & \multicolumn{1}{c|}{84.1} & 86.4 & \multicolumn{1}{c|}{86.7} & \multicolumn{1}{c|}{82.3} & 84.5 \\  
\multicolumn{1}{|c|}{} & \textbf{\name-\docl} & \multicolumn{1}{c|}{\hlg{90.9}} & \multicolumn{1}{c|}{\hlg{89.9}} & \hlg{90.4} & \multicolumn{1}{c|}{\hlg{89.6}} & \multicolumn{1}{c|}{\hlg{88.8}} & \hlg{89.2} \\ \hline
\end{tabular}
\label{tab:perf-impr}
\end{table}

\begin{table}[]
\centering
\caption{Comparative evaluation of \mtd{} performance on the \ctdar{} dataset, comprising English and Chinese documents representing two languages across two distinct scripts reported with an IoU 0.7, and the \dataset{} dataset, our proposed multi-script benchmark featuring diverse languages and document types with an IoU 0.5; best-performing methods (\name) are highlighted.}
\setlength{\tabcolsep}{4.8pt}
\renewcommand{\arraystretch}{1.1}
\centering
\begin{tabular}{|c|c|ccc|ccc|}
\hline
\multirow{2}{*}{\textbf{Baseline}} & \multirow{2}{*}{\textbf{Approach}} & \multicolumn{3}{c|}{\textbf{CTDAR}} & \multicolumn{3}{c|}{\textbf{MANDALA}} \\ \cline{3-8} 
 &  & \multicolumn{1}{c|}{\textbf{P}} & \multicolumn{1}{c|}{\textbf{R}} & \textbf{F} & \multicolumn{1}{c|}{\textbf{P}} & \multicolumn{1}{c|}{\textbf{R}} & \textbf{F} \\ \hline
\multirow{2}{*}{\begin{tabular}[c]{@{}c@{}}\textbf{Multimodal} \\ \textbf{Generative}\end{tabular}} & Qwen-2.5-VL-7B \cite{qwen-vl} & \multicolumn{1}{c|}{69.9} & \multicolumn{1}{c|}{60.8} & 65.0 & \multicolumn{1}{c|}{72.8} & \multicolumn{1}{c|}{64.2} & 68.2 \\  
 & UDOP \cite{udop} & \multicolumn{1}{c|}{17.9} & \multicolumn{1}{c|}{17.5} & 17.7 & \multicolumn{1}{c|}{25.4} & \multicolumn{1}{c|}{23.6} & 24.4 \\ \hline
\multirow{3}{*}{\begin{tabular}[c]{@{}c@{}}\textbf{Pre-reported} \\ \textbf{Scores}\end{tabular}} & DiT-L \cite{dit} & \multicolumn{1}{c|}{-} & \multicolumn{1}{c|}{-} & 97.6 & \multicolumn{1}{c|}{-} & \multicolumn{1}{c|}{-} & - \\  
 & NLPR-PAL \cite{icdar19} & \multicolumn{1}{c|}{96.2} & \multicolumn{1}{c|}{96.9} & 96.6 & \multicolumn{1}{c|}{-} & \multicolumn{1}{c|}{-} & - \\  
 & CornerNet+FRCN \cite{cornernet} & \multicolumn{1}{c|}{98.2} & \multicolumn{1}{c|}{93.9} & 96.0 & \multicolumn{1}{c|}{-} & \multicolumn{1}{c|}{-} & - \\ \hline
\multirow{5}{*}{\begin{tabular}[c]{@{}c@{}}\textbf{Traditional} \\ \textbf{Object} \\ \textbf{Detection} \\ \textbf{Architectures}\end{tabular}} & Faster RCNN X-101 \cite{layoutparser} & \multicolumn{1}{c|}{75.6} & \multicolumn{1}{c|}{72.6} & 74.1 & \multicolumn{1}{c|}{70.7} & \multicolumn{1}{c|}{66.1} & 68.3 \\  
 & Mask RCNN X-101 \cite{layoutparser} & \multicolumn{1}{c|}{78.7} & \multicolumn{1}{c|}{75.4} & 77.0 & \multicolumn{1}{c|}{34.8} & \multicolumn{1}{c|}{32.0} & 33.3 \\  
 & \yolov-PTN \cite{yolo-pub} & \multicolumn{1}{c|}{86.0} & \multicolumn{1}{c|}{81.6} & 83.8 & \multicolumn{1}{c|}{55.0} & \multicolumn{1}{c|}{49.8} & 52.5 \\  
 & \tatr-PTN \cite{tatr-pub-1m} & \multicolumn{1}{c|}{76.2} & \multicolumn{1}{c|}{69.9} & 72.9 & \multicolumn{1}{c|}{55.6} & \multicolumn{1}{c|}{49.2} & 52.2 \\  
 & \docl-PTN \cite{doclayout} & \multicolumn{1}{c|}{82.9} & \multicolumn{1}{c|}{80.8} & 81.8 & \multicolumn{1}{c|}{71.5} & \multicolumn{1}{c|}{67.9} & 69.6 \\ \hline
\multirow{3}{*}{\begin{tabular}[c]{@{}c@{}}\textbf{Fine-tuned} \\ \textbf{on Unmasked} \\ \textbf{Images}\end{tabular}} & \yolov-FTN & \multicolumn{1}{c|}{97.8} & \multicolumn{1}{c|}{98.2} & 98.0 & \multicolumn{1}{c|}{86.4} & \multicolumn{1}{c|}{83.5} & 84.9 \\  
 & \tatr-FTN & \multicolumn{1}{c|}{97.9} & \multicolumn{1}{c|}{96.8} & 97.3 & \multicolumn{1}{c|}{88.4} & \multicolumn{1}{c|}{83.2} & 85.7 \\  
 & \docl-FTN & \multicolumn{1}{c|}{\hlg{98.6}} & \multicolumn{1}{c|}{95.3} & 96.9 & \multicolumn{1}{c|}{81.7} & \multicolumn{1}{c|}{76.2} & 78.8 \\ \hline
\multirow{3}{*}{\textbf{\begin{tabular}[c]{@{}c@{}}Fine-tuned \\ on Masked \\ Images (Ours)\end{tabular}}} & \textbf{\name-\yolo} & \multicolumn{1}{c|}{97.4} & \multicolumn{1}{c|}{\hlg{97.8}} & \hlg{97.6} & \multicolumn{1}{c|}{87.6} & \multicolumn{1}{c|}{86.9} & 87.3 \\  
 & \textbf{\name-\tatr} & \multicolumn{1}{c|}{95.5} & \multicolumn{1}{c|}{93.7} & 94.6 & \multicolumn{1}{c|}{88.9} & \multicolumn{1}{c|}{84.1} & 86.4 \\  
 & \textbf{\name-\docl} & \multicolumn{1}{c|}{97.5} & \multicolumn{1}{c|}{97.1} & 97.3 & \multicolumn{1}{c|}{\hlg{90.9}} & \multicolumn{1}{c|}{\hlg{89.9}} & \hlg{90.4} \\ \hline
\end{tabular}
\label{tab:mtd2}
\end{table}

\subsection{Comparison with State-of-the-Art}
\label{sec:comparison}
Table~\ref{tab:mtd2} shows that \name{} consistently outperforms multimodal generative models and traditional object detection architectures across both \ctdar{} and \dataset{} datasets. In addition to their inferior detection accuracy, VLM-based approaches (Qwen2.5-VL \cite{qwen-vl}, UDOP \cite{udop}) are substantially slower at inference due to autoregressive decoding and high computational overhead, making them computationally expensive and less suited for large-scale deployment. In contrast, \name{} leverages efficient object detection backbones, delivering significantly better accuracy while retaining fast, scalable inference. The absolute 20.8\% F1-score gain achieved by \name-\docl{} over the strongest \docl-PTN baseline further highlights that \name{} improvements stem from superior structural learning rather than increased model complexity. As shown in Table \ref{tab:mtd3}, \name{} ranks second on the IIIT-AR-13K \cite{iiit-ar} and at par on the challenging TNCR \cite{tncr} dataset. As shown in Table~\ref{tab:mtd4}, \name{} matches or surpasses strong task-specific methods on English benchmarks, despite being designed for multi-script generalization. \name-\tatr{} achieves the highest F1-score on ICDAR 2013 \cite{icdar13}, while \name-\yolo{} remains competitive with specialized baselines on PubTables \cite{tatr-pub-1m}. These results confirm that \name{} does not sacrifice performance on English-only datasets while delivering substantial gains in multi-script settings. Table \ref{tab:lang-wise} displays the language-wise performance on \dataset{} using \name-\docl{} at an IoU threshold of 0.5.

\begin{table}[]
\centering
\caption{Comparative overview of \mtd{} performance on the IIIT-AR-13K dataset, comprising documents in four languages written in a single script, and the TNCR dataset, featuring complex real-world table layouts. Results are reported in terms of F1-score (IoU 0.5); best-performing methods are highlighted.}
\setlength{\tabcolsep}{6pt}
\renewcommand{\arraystretch}{1.1}
\begin{tabular}{|c|cc|cc|}
\hline
\textbf{Dataset} & \multicolumn{2}{c|}{\textbf{IIIT-AR-13K}} & \multicolumn{2}{c|}{\textbf{TNCR}} \\ \hline
\textbf{Evaluation} & \multicolumn{1}{c|}{\textbf{Approach}} & \textbf{F} & \multicolumn{1}{c|}{\textbf{Approach}} & \textbf{F} \\ \hline
\multirow{2}{*}{\textbf{\begin{tabular}[c]{@{}c@{}}Prereported \\ Scores\end{tabular}}} & \multicolumn{1}{c|}{Faster RCNN \cite{iiit-ar}} & 93.7 & \multicolumn{1}{c|}{Cascade RPN \cite{tncr}} & \hlg{94.1} \\  
 & \multicolumn{1}{c|}{Mask RCNN \cite{iiit-ar}} & \hlg{97.1} & \multicolumn{1}{c|}{Hybrid Task Cascade \cite{tncr}} & 93.6 \\ \hline
\multirow{6}{*}{\textbf{\begin{tabular}[c]{@{}c@{}}Fine-tuned \\ On Unmasked \\ Images\end{tabular}}} & \multicolumn{1}{c|}{\yolov-PTN} & 71.7 & \multicolumn{1}{c|}{\yolov-PTN} & 69.4 \\  
 & \multicolumn{1}{c|}{\tatr-PTN} & 71.0 & \multicolumn{1}{c|}{\tatr-PTN} & 76.0 \\  
 & \multicolumn{1}{c|}{\docl-PTN} & 81.9 & \multicolumn{1}{c|}{\docl-PTN} & 79.9 \\  
 & \multicolumn{1}{c|}{\textbf{\yolov-FTN}} & 90.8 & \multicolumn{1}{c|}{\textbf{\yolov-FTN}} & 86.2 \\  
 & \multicolumn{1}{c|}{\textbf{\tatr-FTN}} & 93.6 & \multicolumn{1}{c|}{\textbf{\tatr-FTN}} & 89.4 \\  
 & \multicolumn{1}{c|}{\textbf{\docl-FTN}} & 90.5 & \multicolumn{1}{c|}{\textbf{\docl-FTN}} & 80.9 \\ \hline
\multirow{3}{*}{\textbf{Ours}} & \multicolumn{1}{c|}{\textbf{\name-\yolo}} & 93.1 & \multicolumn{1}{c|}{\textbf{\name-\yolo}} & 87.2 \\  
 & \multicolumn{1}{c|}{\textbf{\name-\tatr}} & 93.7 & \multicolumn{1}{c|}{\textbf{\name-\tatr}} & 89.5 \\  
 & \multicolumn{1}{c|}{\textbf{\name-\docl}} & 93.3 & \multicolumn{1}{c|}{\textbf{\name-\docl}} & 85.4 \\ \hline
\end{tabular}
\label{tab:mtd3}
\end{table}

\begin{table}[]
\centering
\caption{Comparative overview of \mtd{} performance on ICDAR 2013 (F1@IoU 0.5) and \pubm{} (AP50) datasets respectively, with best scores highlighted.}
\setlength{\tabcolsep}{6pt}
\renewcommand{\arraystretch}{1}
\begin{tabular}{|c|c|c|cc|}
\hline
\textbf{Dataset} &  \multicolumn{2}{c|}{\textbf{ICDAR 2013}} & \multicolumn{2}{c|}{\textbf{PubTables}} \\ \hline
\textbf{Evaluation} & \textbf{Approach} & \textbf{F} & \multicolumn{1}{c|}{\textbf{Approach}} & \textbf{AP50} \\ \hline
\multirow{5}{*}{Prereported} & DeepDeSRT \cite{deepdesrt} & 96.8 & \multicolumn{1}{c|}{FRCNN \cite{tatr-pub-1m}} & 98.5 \\  
 & TableBank \cite{tablebank} & 96.3 & \multicolumn{1}{c|}{SAM-DETR \cite{sam-detr}} & 94.8 \\  
 & CDec-Net \cite{cdecnet} & 96.8 & \multicolumn{1}{c|}{ClusterTabNet \cite{clustertabnet}} & 99.0 \\  
 & TableNet \cite{TableNet} & 96.6 & \multicolumn{1}{c|}{YOLOv5 \cite{yolo-ptn}} & \hlg{99.5} \\  
 & SAL-CNN \cite{sal-cnn} & 97.8 & \multicolumn{1}{c|}{TATR \cite{tatr-pub-1m}} & \hlg{99.5} \\ \hline
\multirow{3}{*}{\textbf{Ours}} & \textbf{\name-\yolo} & 97.9 & \multicolumn{1}{c|}{\textbf{\name-\yolo}} & 98.1 \\  
 & \textbf{\name-\tatr} & \hlg{99.0} & \multicolumn{1}{c|}{\textbf{\name-\tatr}} & 99.0 \\  
 & \textbf{\name-\docl} & 96.6 & \multicolumn{1}{c|}{\textbf{\name-\docl}} & 95.8 \\ \hline
\end{tabular}
\label{tab:mtd4}
\end{table}

\begin{table}
\centering
\caption{Language-wise table detection performance on \dataset{} using \name-\docl{} at an IoU threshold of 0.5, yielding an overall F1-score of 90.4\%.}
\setlength{\tabcolsep}{2.5pt}
\renewcommand{\arraystretch}{1}
\begin{tabular}{|c|c|c|c|c|c|c|c|c|c|c|c|}
\hline
\textbf{Lang} & \textbf{P} & \textbf{R} & \textbf{F} &
\textbf{Lang} & \textbf{P} & \textbf{R} & \textbf{F} &
\textbf{Lang} & \textbf{P} & \textbf{R} & \textbf{F} \\
\hline
\textbf{Bengali}   & 91.3 & 92.1 & 91.7 & \textbf{Japanese}  & 91.6 & 88.5 & 90.0 & \textbf{Persian}   & 92.4 & 89.0 & 90.6 \\
\textbf{Greek}     & 94.7 & 91.8 & 93.3 & \textbf{Kannada}   & 89.7 & 93.4 & 91.5 & \textbf{Punjabi}   & 92.0 & 93.6 & 92.8 \\
\textbf{Gujarati}  & 92.8 & 95.3 & 94.0 & \textbf{Khmer}     & 96.4 & 91.5 & 93.9 & \textbf{Russian}   & 94.5 & 86.3 & 90.2 \\
\textbf{Hindi}     & 86.1 & 85.6 & 85.9 & \textbf{Malayalam} & 93.8 & 95.7 & 94.8 & \textbf{Tamil}     & 75.6 & 81.7 & 78.6 \\
\textbf{Marathi}   & 98.9 & 97.4 & 98.1 & \textbf{Odia}      & 82.5 & 82.7 & 82.6 & \textbf{Telugu}    & 86.2 & 90.9 & 88.5 \\
\textbf{Thai}      & 95.4 & 95.9 & 95.6 & \textbf{Ukrainian} & 88.3 & 76.5 & 82.0 & \textbf{Urdu}      & 86.3 & 88.6 & 87.4 \\
\hline
\end{tabular}
\label{tab:lang-wise}
\end{table}

\subsection{Ablation Study}
\label{sec:ablation}
This study validates the effectiveness of \name{}’s strategy of fine-tuning on masked images and inferring on original images for \mtd{}, using \docl-PTN. All experiments use $2,000$ training and $183$ validation images (Section~\ref{sec:datasets}), with fine-tuning performed separately on original and masked images. Each fine-tuned model was evaluated under two inference conditions: original (unmasked) images and masked images as input. As shown in Table \ref{tab:ablation}, the \textit{Masked-Original} paradigm outperformed the others, leveraging pre-training on original images while fine-tuning on masked ones, thereby improving generalization without requiring multi-script, labeled data. The performance gap for \mtd{} is evident when the pretrained \docl{} model is used for direct inference.
In contrast, since the \textit{Original-Masked} method performed the worst, and \textit{Masked-Masked} introduces an unnecessary masking step at inference, we do not advocate for those approaches. The same trend was also observed for other models like \tatr{} and \yolov{} across all IoU thresholds. This confirms that masking is most effective as a regularizer during fine-tuning, rather than as an inference-time transformation, allowing \name{} to retain the learned structural priors from the original images while mitigating script bias.

\begin{table*}[h!]
\centering
\renewcommand{\arraystretch}{1}
\setlength{\tabcolsep}{6pt}
\caption{Ablation on \ctdar{} and \dataset{} evaluates four fine-tuning–inference combinations using \textbf{\docl} at IoU 0.5; best scores highlighted.}
\begin{tabular}{|c|c|ccc|ccc|}
\hline
\multirow{2}{*}{\textbf{\docl{} Model}} & \multirow{2}{*}{\textbf{Inferred on}} & \multicolumn{3}{c|}{\textbf{\ctdar}} & \multicolumn{3}{c|}{\textbf{MANDALA}} \\ \cline{3-8} 
 &  & \multicolumn{1}{c|}{\textbf{P}} & \multicolumn{1}{c|}{\textbf{R}} & \textbf{F} & \multicolumn{1}{c|}{\textbf{P}} & \multicolumn{1}{c|}{\textbf{R}} & \textbf{F} \\ \hline
\multirow{2}{*}{Pretrained} & Original & \multicolumn{1}{c|}{86.1} & \multicolumn{1}{c|}{83.7} & 84.9 & \multicolumn{1}{c|}{71.5} & \multicolumn{1}{c|}{67.9} & 69.6 \\  
 & Masked & \multicolumn{1}{c|}{41.3} & \multicolumn{1}{c|}{40.1} & 40.7 & \multicolumn{1}{c|}{59.5} & \multicolumn{1}{c|}{57.1} & 58.3 \\ \hline
\multirow{2}{*}{Fine-tuned - Original} & Original & \multicolumn{1}{c|}{\hlg{98.6}} & \multicolumn{1}{c|}{95.3} & 96.9 & \multicolumn{1}{c|}{81.7} & \multicolumn{1}{c|}{76.2} & 78.8 \\  
 & Masked & \multicolumn{1}{c|}{67.3} & \multicolumn{1}{c|}{62.2} & 64.6 & \multicolumn{1}{c|}{67.2} & \multicolumn{1}{c|}{61.7} & 64.3 \\ \hline
\multirow{2}{*}{Fine-tuned - Masked} & Original & \multicolumn{1}{c|}{97.7} & \multicolumn{1}{c|}{\hlg{97.5}} & 97.6 & \multicolumn{1}{c|}{\hlg{90.9}} & \multicolumn{1}{c|}{\hlg{89.9}} & \hlg{90.4} \\  
 & Masked & \multicolumn{1}{c|}{98.3} & \multicolumn{1}{c|}{97.2} & \hlg{97.7} & \multicolumn{1}{c|}{85.9} & \multicolumn{1}{c|}{83.1} & 84.5 \\ \hline
\end{tabular}
\label{tab:ablation}
\end{table*}


\section{Conclusion}
\label{sec:future}
We present \name{}, a model-agnostic masking-based fine-tuning paradigm that substantially improves \mtd{} by encouraging reliance on script-independent structural cues. Extensive experiments reveal that fine-tuning on masked images followed by inference on original images is the most effective configuration. \name{} yields consistent cross-script gains irrespective of pretraining dataset, as well as without requiring any multilingual (non-Latin) training data. Importantly, these improvements are achieved without degrading performance on Latin-script documents, demonstrating that \name{} enhances generalization. The effectiveness of \name{} is validated across multiple architectures, including \yolov{}, \tatr{}, and \docl{}, underscoring its general applicability. Complementing this method, we also introduce \dataset{}, a challenging multi-script benchmark spanning 18 languages and 15 scripts, designed to assess cross-script \td{} robustness. Together, \name{} and \dataset{} establish a strong foundation for scalable, script-agnostic table detection and provide a direction for advancing multi-script document understanding.

\section{Limitations and Future Work}
While \name{} focuses on table detection, it does not explicitly address table structure recognition or nested tables, which remain challenging in complex documents. The current masking strategy relies on fixed thresholds and binarization, which may limit performance on low-quality or visually degraded inputs. Future work will explore adaptive masking strategies and richer intermediate representations, such as grayscale, using dynamic thresholds or blurred masks. We also plan to extend this framework to script-agnostic table structure recognition and full-page multiclass layout analysis, enabling unified modeling of rows, columns, spanning cells, and heterogeneous layout elements such as figures, equations, titles, and text blocks within multi-script documents.

\subsubsection*{Acknowledgement}
We gratefully acknowledge the Government of India for its support through MeitY and DST, via their respective initiatives, Bhashini and BharatGen, which were instrumental in enabling this work.




\bibliographystyle{splncs04}
\bibliography{bibliography}

\end{document}